%% file: acl_latex.tex
\pdfoutput=1

\documentclass[11pt]{article}

\usepackage[preprint]{acl}

\usepackage{times}
\usepackage{latexsym}
\usepackage{microtype}
\usepackage{graphicx}
\usepackage{caption}
\usepackage{subcaption}
\usepackage{booktabs} %
\usepackage{tabularx}
\usepackage{amsmath}
\usepackage{xcolor}
\definecolor{TODO}{rgb}{1,0,0} %
\usepackage{amssymb}
\usepackage{mathtools}
\usepackage{amsthm}
\usepackage{lipsum}  
\usepackage{listings}
\usepackage{CJKutf8}
\usepackage{xspace}
\usepackage{multirow}
\usepackage[framemethod=TikZ]{mdframed}
\usepackage{tabu}
\usepackage{longtable}
\usepackage{alphalph}
\usepackage{bm}
\usepackage{tcolorbox}
\usepackage{pifont}
\usepackage{fontawesome5} %
\usepackage[moderate]{savetrees}

\usepackage[T1]{fontenc}

\usepackage[utf8]{inputenc}

\usepackage{microtype}

\usepackage{inconsolata}

\usepackage{graphicx}
\usepackage{hyperref}
\hypersetup{
    colorlinks=,
    linkcolor=,
    filecolor=,      
    urlcolor=,
    }

\newcolumntype{C}{>{\centering\arraybackslash}X}

\definecolor{promptbg}{RGB}{243,244,246}   %
\definecolor{promptborder}{RGB}{56,189,248} %

\tcbset{
  promptbox/.style={
    colback=promptbg,
    colframe=promptborder,
    boxrule=0.8pt,
    arc=5pt,
    left=4pt,
    right=4pt,
    top=4pt,
    bottom=4pt,
    fonttitle=\bfseries,
    before skip=10pt, after skip=10pt,
    boxsep=4pt,
    width=\linewidth,
    fontupper=\ttfamily\small,
  }
}

\title{\textbf{Puzzled by Puzzles:} When Vision-Language Models Can't Take a Hint}

\author{\textbf{Heekyung Lee}\textsuperscript{1,2}~\quad\textbf{Jiaxin Ge}\textsuperscript{2}~\quad \textbf{Tsung-Han Wu}\textsuperscript{2}~\quad\textbf{Minwoo Kang}\textsuperscript{2}\\
\textbf{Trevor Darrell}\textsuperscript{2}~\quad
\textbf{David M. Chan}\textsuperscript{2} \\[8pt]
\textsuperscript{1}POSTECH~\quad \textsuperscript{2}University of California, Berkeley
}

\begin{document}

\maketitle

\input{sections/0_abstract}
\input{sections/1_introduction}

\input{sections/2_methods}
\input{sections/3_results}
\input{sections/4_conclusion}

\section*{Acknowledgements}

We would like to express our sincere gratitude to Lisa Dunlap, XuDong Wang, Konpat Preechakul, Baifeng Shi, and Stephanie Murphy for their invaluable assistance in paper review, ideation, and puzzle solving. As part of their affiliation with UC Berkeley, the authors were supported in part by the National Science Foundation, the Ford Foundation, and/or the Berkeley Artificial Intelligence Research (BAIR) Industrial Alliance program. This material is based upon work supported by the Defense Advanced Research Projects Agency (DARPA), the Army Contracting Command-Aberdeen Proving Grounds (ACC-APG), and the Air Force Research Laboratory. The views, opinions, and/or findings expressed are those of the authors and should not be interpreted as representing the official views or policies of any supporting entity, including the AFRL, ACC-APG, the Department of Defense or the U.S. Government.

\clearpage

\bibliography{custom}

\clearpage
\appendix

\renewcommand{\theequation}{\thesection.\arabic{equation}}
\renewcommand{\thefigure}{\thesection.\arabic{figure}}
\renewcommand{\thetable}{\thesection.\arabic{table}}

\makeatletter
\@addtoreset{equation}{section}
\@addtoreset{figure}{section}
\@addtoreset{table}{section}
\makeatother

\section*{Appendix}
\label{sec:appendix}

\input{sections/5_appendix}

\end{document}

%% file: sections/0_abstract.tex
\begin{abstract}
   Rebus puzzles, visual riddles that encode language through imagery, spatial arrangement, and symbolic substitution, pose a unique challenge to current vision-language models (VLMs). Unlike traditional image captioning or question answering tasks, rebus solving requires multi-modal abstraction, symbolic reasoning, and a grasp of cultural, phonetic and linguistic puns. In this paper, we investigate the capacity of contemporary VLMs to interpret and solve rebus puzzles by constructing a hand-generated and annotated benchmark of diverse English-language rebus puzzles, ranging from simple pictographic substitutions to spatially-dependent cues ("head" over "heels"). We analyze how different VLMs perform, and our findings reveal that while VLMs exhibit some surprising capabilities in decoding simple visual clues, they struggle significantly with tasks requiring abstract reasoning, lateral thinking, and understanding visual metaphors.
\end{abstract}

\begin{center}
{
\small
{\faGlobe} \href{https://visual-puzzles.github.io}{\textcolor{magenta}{\textbf{Project Page}}} \quad
\faGithub\ \href{https://visual-puzzles.github.io}{\textcolor{magenta}{\textbf{Code}}} \ \hspace{0.3em}
\raisebox{-0.2em}{\includegraphics[height=1.1em]{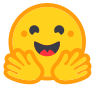}}\
\href{https://huggingface.co/datasets/Kyunnilee/visual-puzzles}{\textcolor{magenta}{\textbf{Datasets}}}
}
\end{center}

%% file: sections/1_introduction.tex
\section{Introduction \& Background}

Vision-language models (VLMs) have demonstrated rapid progress across a range of multi-modal tasks, from image captioning and visual question answering to more structured reasoning benchmarks. However, there remain fundamental gaps in our understanding of these models' ability to perform both abstract and lateral reasoning - capabilities which are critical to human cognition. This gap in understanding is, in part, due to the fact that standard evaluations for vision-language modeling tend to reward direct visual-text alignment or rote recall, leaving open whether VLMs can generalize to compositional, symbolic, and context-dependent inferences. 

Towards solving such issues with standard benchmarking, ``puzzle-based'' evaluations have been explored as an alternative test for higher-order AI reasoning. In VLMs, early studies tested models on games such as Iconary \cite{clark-etal-2021-iconary} and IQ-style challenges \cite{huang2023language}, with recent benchmarks such as VGRP-Bench \cite{ren2025vgrp} using visual grid logic puzzles to demonstrate that multi-modal language models struggle with systematic deduction and TruthQuest \cite{mondorf2024liar} using ``Knights and Knaves'' logic puzzles to test LLM suppositional reasoning. In textual riddles, benchmarks like RiddleSense \cite{lin2021riddlesense} test basic reasoning, while BRAINTEASER \cite{jiang2023brainteaser} and LatEval \cite{huang2023lateval} focus on lateral thinking, showing models often default to incorrect commonsense answers or struggle with interactive inquiry. Bongard in Wonderland \cite{dai2024gpt} tests puzzles that require basic human-like abilities, revealing that VLMs struggle even to detect basic visual concepts and perform abstract reasoning.

\begin{figure}
    \centering
    \includegraphics[width=\linewidth]{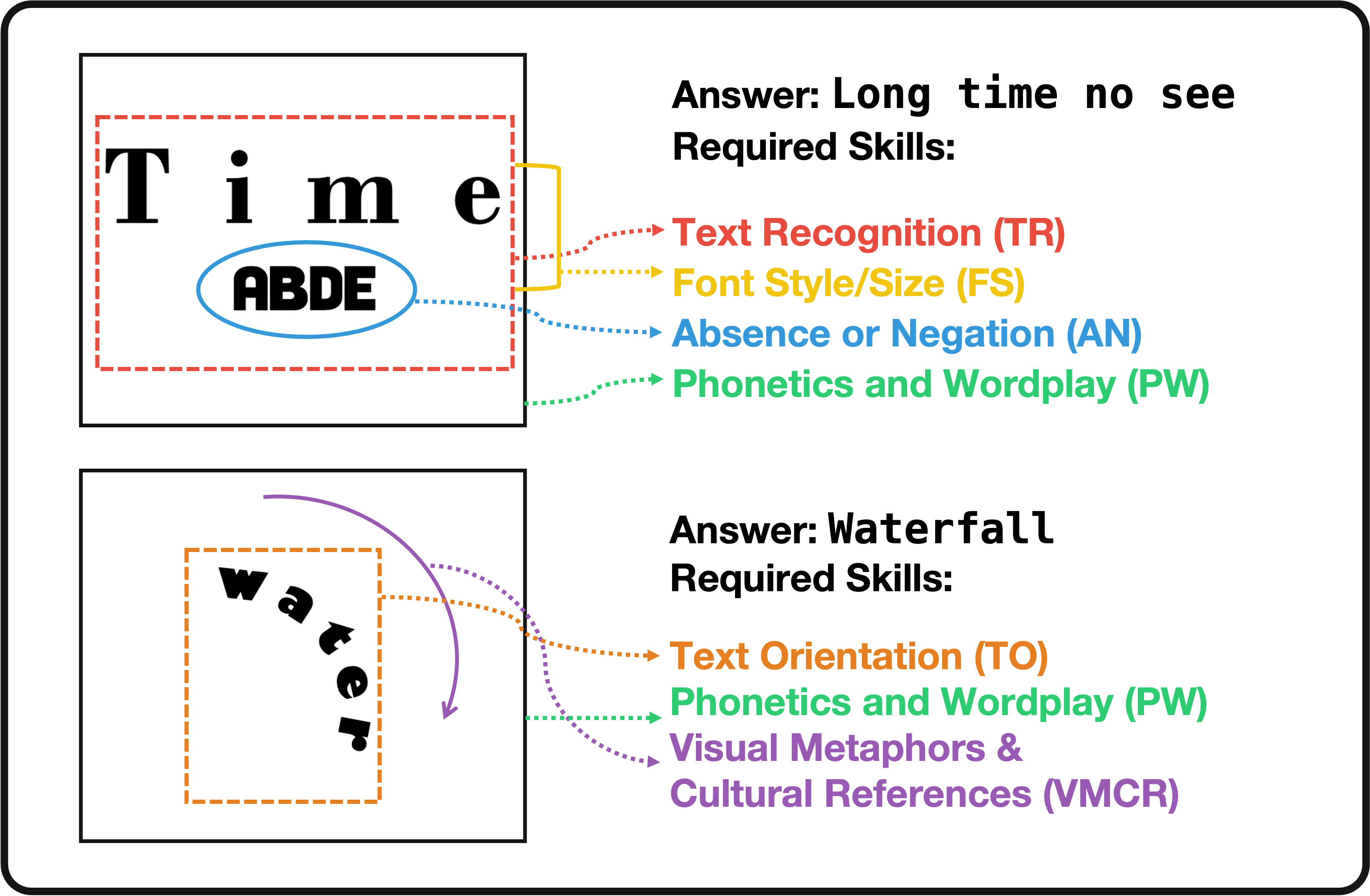}
    \caption{\textbf{Rebus Puzzles}: Two example rebus puzzles along with the cognitive skills required to solve them In this short paper, we use a set of 432 hand-created and annotated rebus puzzles to map the capabilities and limitations of VLMs.}
    \label{fig:teaser}
\end{figure}

One particularly interesting puzzle type, ``rebus puzzles,'' is a form of visual wordplay that encodes phrases, idioms, or concepts through combinations of images, letters, symbols, and spatial arrangements. Solving a rebus puzzle requires more than image recognition or surface-level language modeling: it calls for the integration of visual cues with symbolic abstraction, phonetic manipulation, and a sensitivity to cultural and linguistic context. For example, the word ``WATER'' in a curved downward shape (\autoref{fig:teaser}) evokes the phrase ``Waterfall.'' These puzzles can be simple to humans (particularly expert puzzle solvers), but deceptively rich in the kinds of cognitive processes they require.  Recently, \citet{gritsevskiy2024rebus} introduced a benchmark using drawn/digital images rebus puzzles that require broad world knowledge.  While an exciting first step, the benchmark includes a heavy skew towards particular categories, for example, 28\% of the dataset is Massachusetts towns. Following this work, \citet{kraaijveld2025columbus} developed an automatic generator for several types of rebus-like tasks, but evaluated models within a multiple-choice context and tested only a small subset of generative rules.

In this short paper, we use a hand-generated and annotated probe dataset of 432 rebus puzzles, each labeled with both the ground truth solution and detailed cognitive skill categories, to systematically examine the visual and reasoning capabilities of modern vision and language models. We evaluate not only overall model performance and how it compares to human solvers, but also how different forms of context, prompt design, and skill-specific cues influence models' puzzle-solving ability. To understand this accuracy further, we additionally examine the robustness of models to prompt variations, their ability to self-correct when given feedback, and how performance varies across different modalities and reasoning types. To our knowledge, this is the first work to study reasoning models on open-ended rebus tasks (A detailed comparison with prior work is provided in \autoref{app:rebus}). Overall, while we find that models, particularly reasoning models, exhibit partial competency on rebus puzzles, even the strongest models lag substantially behind human solvers on puzzles that require deeper levels of lateral thinking, primarily driven by weaknesses in \textit{reasoning capability} rather than \textit{perception and understanding}.

%% file: sections/2_methods.tex
\section{Dataset \& Metrics}
In this short paper, we construct a human-crafted probe dataset of rebus puzzles that can challenge VLMs to reason effectively about visual-linguistic information. We collected puzzle images via an online resource\footnote{\href{https://eslvault.com/free-printable-rebus-puzzles/}{ESL Vault: Rebus Puzzles}} for rebus puzzles. The authors manually inspected, filtered, and corrected each puzzle image to ensure accuracy and consistent resolution throughout the dataset with a minimum of two annotations per puzzle. 

Alongside the puzzle images and answers, the authors further annotated each puzzle with a set of visual-linguistic skills required to solve that puzzle. Puzzles were organized into 11 distinct cognitive categories: Absence or Negation (AN), Text Orientation (TO), Quantitative or Mathematical Reasoning (QMR), Visual Metaphors and Cultural References (VMCR), Symbolic Substitution (SS), Font Style/Size (FS), Letter and Word Manipulation (LWM), Phonetics and Wordplay (PW), Spatial and Positional Reasoning (SPR), Image Recognition (IR) and Text Recognition (TR). A detailed description of each of the categories is given in \autoref{table:skill_distribution}.

To evaluate the performance of VLMs, we adopt two evaluation methods: ``Naive Matching'' and ``LLM-Judged'' evaluation. The Naive Matching method evaluates responses based on exact string matching with the ground-truth answers, while the LLM-Judged evaluation method uses Prompted LLMs (\texttt{gpt-4o}, \texttt{Qwen3-8B}) to compare ground truth and candidate answers directly (see \autoref{app:evaluation} for details). 

%% file: sections/3_results.tex
\input{table/result_table}

\input{table/accuracy_per_skill}

\section{Results \& Discussion}
\label{sec:results}

Overall performance on the probe dataset is given in \autoref{table:rebus_results}. We can see that models have a wide range of performance - while closed source reasoning models such as \texttt{GPT-5}, \texttt{o3}, and \texttt{o4-mini} perform relatively well, especially compared to human non-expert/non-native English speaking solvers, open-source reasoning models as non-reasoning models struggle to solve the tasks. Compared to expert solvers, however, there remains a significant model accuracy gap. 

Skill-specific results are given in \autoref{table:skill_performance}. While models are surprisingly competent in solving symbolic manipulation (SS/SPR) and quantitative reasoning samples (QMR, likely to to a focus on math-specific benchmarks during training), they have significant gaps in abstract reasoning/lateral thinking, particularly in recognizing the absence of objects (AN), and in solving visual metaphors (VMCR). Indeed, the general gap between performance in symbolic substitution (SS) vs. phonetics and wordplay (PW) suggests that VLMs are better at learning and applying these more direct, almost rule-based symbolic transformations. 

The gap between spatial and positional reasoning (SPR) and letter and word manipulation (LWM) in most models is similarly interesting: In models that perform decently on SPR but struggle more with LWM, we see that they can understand the layout of elements but fail when the puzzle demands a subsequent, often abstract, manipulation or reinterpretation of the textual components themselves. This general distinction implies a disconnect between understanding spatial configuration and then applying complex linguistic or symbolic logic to the content within that configuration.

Notable as well, is the interplay in performance between text recognition (TR), font-style/size (FS) and text orientation (TO). Generally high TR paired with relatively high TO (for the best models) suggests that models can read text even when its orientation is unusual. However, somewhat lower scores in FS for these models compared to their TR indicate that while they can read the text, they are not always effectively using cues from font style or size to derive the solution. 

In general, from this exploration we confirm that VLMs (and even reasoning VLMs) are stronger on tasks that are more perceptual or involve more direct, learned mappings and weaker on tasks requiring deeper levels of abstraction. 

\paragraph{In-Context Learning}
A larger question remains as to if models understand the concept of a ``rebus'' puzzle from the prompt alone. Following \citet{brown2020language}, we phrase the task as an in-context learning problem to remove this prompt-specific bias (See \autoref{app:in-context learning} for details). \autoref{table:icl} shows the performance of providing a single in-context example consisting of an image, an answer, and reasoning. Model performance is largely the same, with Qwen2.5-VL showing minor gains, suggesting that performance is not prompt-limited, and instead, is inherent to the underlying VLM. It's also worth noting that ICL is unlikely to help stronger models with more capable reasoning, as each visual puzzle requires additional lateral thinking, and examples often do not allow for such generalization. Smaller models like Qwen2.5-VL, however, may not understand the ``rebus'' concept,  may benefit from a reasoning example.

\input{table/icl}

\paragraph{Skill-Guided Prompting.} In addition to general puzzle understanding, we also wanted to explore if models lack the capability to understand and apply which cognitive thinking skills are required for each puzzle. \autoref{table:skill_guided} demonstrates the impact of providing models with the required skills necessary to solve each puzzle in the prompt itself (See \autoref{app:skill-prompting} for details). Interestingly, incorporating these skill-related prompts generally leads to minor improvements in model performance, suggesting the existence of an \textit{awareness vs. execution gap}: the primary bottleneck for VLMs in solving complex rebus puzzles may not be a failure to identify what kind of reasoning is needed, but rather a more fundamental limitation in how to execute that reasoning effectively.

\input{table/skills_annotation}

\begin{figure}
    \centering
    \includegraphics[width=\linewidth]{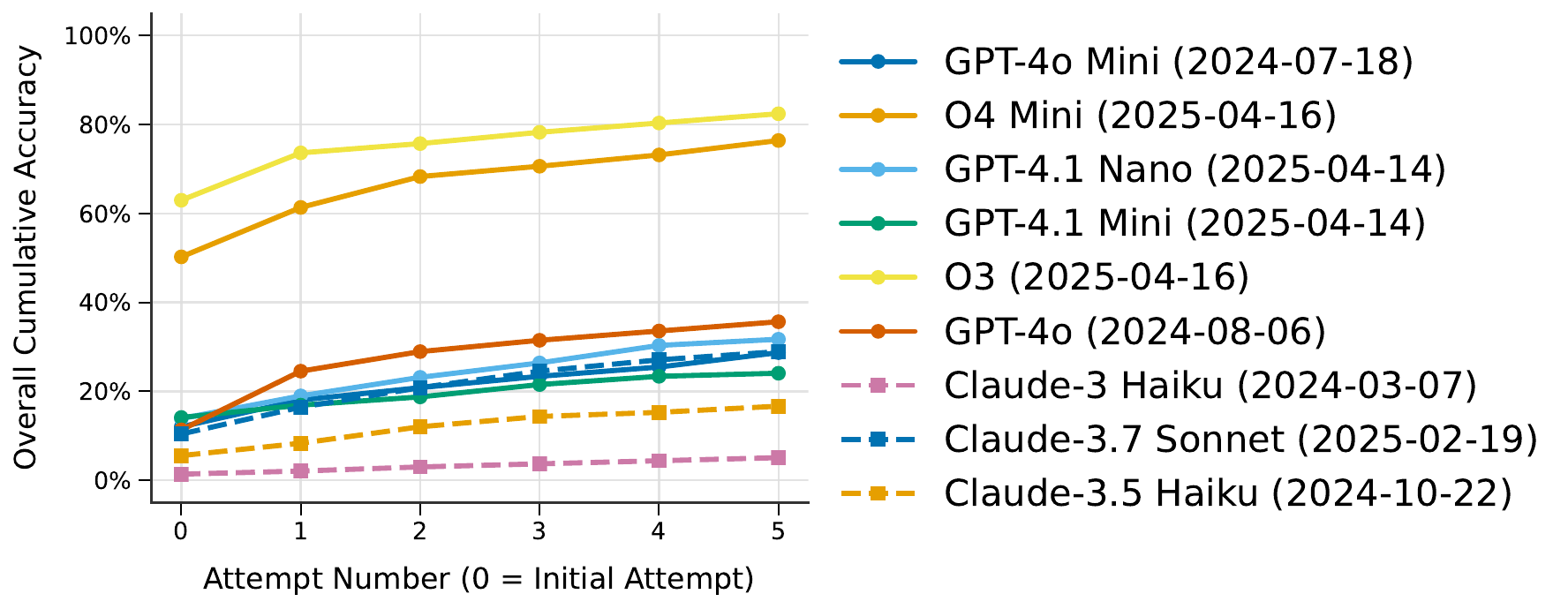}
    \caption{\textbf{Iterative refinement performance.} In general, multiple attempts lead to nominal performance gains.}
    \label{fig:attempts}
\end{figure}

\paragraph{Iterative Refinement.} Arising from a lack of overall performance, we also wondered if perhaps models are locally close to the solution, and could refine the solution given the opportunity. We thus explored an experimental setup in which models were prompted with a ``retry'' prompt if they failed the first time (see \autoref{app:multiple-attempts} for details). The results, given in \autoref{fig:attempts}, show that while models can experience improvement over time, they still reach a performance ceiling after a number of attempts. There is, however, a notable gap between accuracy after 1 and 5 attempts, even in the weakest models, indicating significant potential for iterative refinement or self-correction in models' puzzle-solving processes.

\paragraph{Vision? Or Language?} To further analyze poor VLM performance, we aimed to understand how closely, if at all, model performance was tied to visual perception quality. To test this, we replaced each image with a detailed caption of the image, and used the same puzzle solving prompt, this time conditioned on language alone (see \autoref{app:caption-only}). The results in \autoref{table:captioning} show that models, particularly reasoning models, suffer from a lack of access to the direct visual input.  We hypothesize that the lack of direct visual access is significantly more impactful for reasoning models due to a lack of the models' ability to perform iterative examination of the underlying visual content during the decoding/reasoning process, and it is interesting and exciting future work to explore the extent to which such iterative reasoning processes impact overall downstream performance in VLMs. 

\input{table/captioning_comparison}

\paragraph{Image Retrieval.} To further evaluate the impact of visual reasoning, we explored how underlying visual contrastive models (which are often used as VLM feature extractors) perform in retrieving the correct answers (see \autoref{app:image_retrieval} for details/full results). The results are in  \autoref{tab:vqa_retrieval_split}. Unsurprisingly, architectural design significantly impacts performance, however interestingly, MobileCLIP \cite{vasu2024mobileclip} shows strong results despite its efficiency focus, likely due to underlying data distribution (the DataCompDR dataset). SigLIP~2 \cite{tschannen2025siglip} and TULIP \cite{tang2025tulip} both contain explicit visual reconstruction objectives, which likely lead to their second-best overall accuracies on the task. Model scale and patch size also drive performance, with larger scales (L, GOPT, So400m) and smaller patch sizes (B16 vs B32) generally yielding better metrics, suggesting finer tokenization aids visual feature representation. Notably, for the SigLIP~2 family, increasing resolution beyond 256 did not consistently improve metrics, sometimes showing slight decreases at 384 and 512, confirming that our Rebus puzzles do not require high-fidelity visual encoding and rather require more flexible latent spaces.

\input{table/clip_recall}

%% file: table/result_table.tex
\begin{table}
\scriptsize
\centering
\setlength\tabcolsep{4pt} %
\caption{\textbf{Overall accuracy}\textsubscript{[95\% CI]} on our probe dataset for various VLMs and human scorers.}
\label{table:rebus_results}
\begin{tabularx}{\linewidth}{Xrr}
\toprule
\textbf{Model} & \textbf{Naive Matching} & \textbf{LLM-Judged} \\
\midrule
\texttt{GPT-5}& $60.42_{\textcolor{gray}{[0.556, 0.651]}}$ & $69.44_{\textcolor{gray}{[0.653, 0.736]}}$ \\
\texttt{o3}& $54.56_{\textcolor{gray}{[0.498, 0.590]}}$ & $54.60_{\textcolor{gray}{[0.500, 0.590]}}$ \\
\texttt{o4-mini} & $48.16_{\textcolor{gray}{[0.433, 0.532]}}$ & $55.62_{\textcolor{gray}{[0.509, 0.599]}}$ \\
\texttt{gemini-2.5-pro} & $39.99_{\textcolor{gray}{[0.354, 0.442]}}$ & $47.27_{\textcolor{gray}{[0.424, 0.519]}}$ \\
\texttt{gemini-2.5-flash} & $19.26_{\textcolor{gray}{[0.157, 0.231]}}$ & $24.19_{\textcolor{gray}{[0.204, 0.285]}}$ \\
\texttt{gpt-4o}  & $18.75_{\textcolor{gray}{[0.153, 0.225]}}$ & $26.52_{\textcolor{gray}{[0.225, 0.308]}}$ \\
\texttt{claude-3.7-sonnet} & $10.20_{\textcolor{gray}{[0.076, 0.132]}}$ & $16.31_{\textcolor{gray}{[0.127, 0.199]}}$ \\
\texttt{gemini-2.0-flash} & $8.37_{\textcolor{gray}{[0.058, 0.011]}}$ & $17.25_{\textcolor{gray}{[0.139, 0.206]}}$ \\
\texttt{gpt-4o-mini}  & $8.28_{\textcolor{gray}{[0.058, 0.110]}}$ & $12.28_{\textcolor{gray}{[0.093, 0.155]}}$ \\
\texttt{claude-3.7-haiku} & $6.48_{\textcolor{gray}{[0.042, 0.088]}}$ & $9.03_{\textcolor{gray}{[0.065, 0.120]}}$ \\
\texttt{Molmo-72B-0924}& $1.86_{\textcolor{gray}{[0.007, 0.032]}}$ & $4.43_{\textcolor{gray}{[0.025, 0.065]}}$ \\ 
\texttt{qwen2.5-VL-7B}  & $3.01_{\textcolor{gray}{[0.014, 0.046]}}$ & $3.98_{\textcolor{gray}{[0.021, 0.060]}}$ \\
\texttt{qwen2.5-omni-7B}& $1.88_{\textcolor{gray}{[0.007, 0.032]}}$ & $1.90_{\textcolor{gray}{[0.007, 0.032]}}$ \\
\texttt{Molmo-7B-D-0924}& $1.35_{\textcolor{gray}{[0.004, 0.025]}}$ & $1.85_{\textcolor{gray}{[0.001, 0.030]}}$ \\
\texttt{pixtral-12B} & $1.35_{\textcolor{gray}{[0.005, 0.025]}}$ & $1.39_{\textcolor{gray}{[0.005, 0.255]}}$ \\
\texttt{phi-4}& $0.46_{\textcolor{gray}{[0.000, 0.013]}}$ & $0.96_{\textcolor{gray}{[0.002, 0.019]}}$ \\
\midrule
Human (Non Expert) & $16.4$-$37.23$ & $21.86$-$44.71$ \\
Human (Expert) & $76.4$ & $80.80$ \\
\bottomrule
\end{tabularx}
\end{table}

%% file: table/accuracy_per_skill.tex
\begin{table*}
\scriptsize
\centering
\setlength\tabcolsep{3pt} %
\caption{\textbf{Per-skill model Naive Matching performance} using the base prompt, with (N) the number of samples of each type. Categories: Absence/Negation (AN), Text Orientation (TO), Quantitative Reasoning (QMR), Visual Metaphors/Cultural References (VMCR), Symbolic Substitution (SS), Font Style/Size (FS), Letter and Word Manipulation (LWM), Phonetics/Wordplay (PW), Spatial Reasoning (SPR), Image Recognition (IR), Text Recognition (TR).}
\label{table:skill_performance}
\begin{tabularx}{\linewidth}{lCrrrrrrrrrrr}
\toprule
\textbf{Model} & Reasoning Model & \textbf{AN (12)} & \textbf{TO (14)} & \textbf{QMR (18)} & \textbf{VMCR (23)} & \textbf{SS (35)} & \textbf{FS (43)} & \textbf{LWM (62)} & \textbf{PW (78)} & \textbf{SPR (97)} & \textbf{IR (301)} & \textbf{TR (340)} \\
\midrule
\texttt{qwen2.5-VL-7B} && 0.0 & 0.0 & 0.0 & 0.0 & 5.71 & 0.0 & 1.61 & 1.28 & 3.09 & 1.33 & 2.35 \\
\texttt{Molmo-72B-0924} && 0.0 & 7.14 & 5.56 & 0.0 & 2.86 & 0.0 & 1.61 & 1.28 & 4.12 & 1.0 & 2.35 \\
\texttt{claude-3.5-haiku} && 0.0 & 14.29 & 16.67 & 4.35 & 2.86 & 4.65 & 6.45 & 1.28 & 15.46 & 3.65 & 8.24 \\
\texttt{claude-3.7-sonnet} & \ding{52} & 0.0 & 35.71 & 22.22 & 13.04 & 14.29 & 13.95 & 11.29 & 7.69 & 16.49 & 6.64 & 12.35 \\
\texttt{gemini-2.0-flash} && 8.33 & 28.57 & 22.22 & 8.7 & 8.57 & 6.98 & 16.13 & 5.13 & 13.4 & 4.65 & 9.71 \\
\texttt{gemini-2.5-flash} & \ding{52} & 8.33 & 14.29 & 33.33 & 17.39 & 25.71 & 11.63 & 20.97 & 20.51 & 19.59 & 17.61 & 21.18 \\
\texttt{gpt-4o-mini} && 0.0 & 7.14 & 5.56 & 13.04 & 17.14 & 13.95 & 6.45 & 5.13 & 12.37 & 7.31 & 9.71 \\
\texttt{gpt-4o} && 16.67 & 42.86 & 27.78 & 13.04 & 28.57 & 25.58 & 16.13 & 16.67 & 22.68 & 14.29 & 21.47 \\
\texttt{gemini-2.5-pro} & \ding{52} & \underline{33.33} & 57.14 & 61.11 & 30.43 & 48.57 & 37.21 & 38.71 & 33.33 & 43.30 & 37.87 & 42.94 \\
\texttt{o4-mini}  & \ding{52} & 25.0 & \underline{71.43} & \underline{66.67} & \underline{34.78} & \textbf{65.71} & 32.56 & \underline{43.55} & 42.31 & 43.30 & 45.85 & 50.59 \\
\texttt{o3}  & \ding{52} & \underline{33.33} & \underline{71.43} & \textbf{77.78} & \underline{34.78} & \underline{62.86} & \underline{41.86} & \underline{43.55} & \underline{50.00} & \underline{50.52} & \underline{54.15} & \underline{55.00} \\
\texttt{GPT-5}  &&  \textbf{50.00} & \textbf{78.57} & \textbf{77.78} & \textbf{52.17} & 60.00 & \textbf{51.16} & \textbf{54.84} & \textbf{51.28} & \textbf{62.89} & \textbf{57.81} & \textbf{61.18} \\

\bottomrule
\end{tabularx}
\end{table*}

%% file: table/icl.tex
\begin{table}
\scriptsize
\centering
\setlength\tabcolsep{8pt}
\caption{\textbf{Comparison of models using an ICL example.} Adding an ICL example decreases o4-mini performance, has minimal effect on GPT-4o, and improves Qwen2.5-VL performance.}
\label{table:icl}
\begin{tabularx}{\linewidth}{Xrr}
\toprule
\textbf{Model} & \textbf{Naive} & \textbf{LLM-Judged} \\
\midrule
\texttt{gpt-4o} & 19.03\% ($\uparrow$ 0.28\%) & 25.78\% ($\downarrow$ 0.74\%)\\
\texttt{o4-mini} & 45.83\% ($\downarrow$ 2.33\%)& 53.19\% ($\downarrow$ 2.43\%)\\
\texttt{qwen2.5-VL-7B} & 3.43\% ($\uparrow$ 0.42\%) & 6.70\% ($\uparrow$ 2.72\%)\\
\bottomrule
\end{tabularx}
\end{table}

%% file: table/skills_annotation.tex
\begin{table}
\scriptsize
\centering
\setlength\tabcolsep{8pt}
\caption{\textbf{Comparison of models using a skill guidance example.}}
\label{table:skill_guided}
\begin{tabularx}{\linewidth}{Xrr}
\toprule
\textbf{Model} & \textbf{Naive} & \textbf{LLM-Judged} \\
\midrule
\texttt{gpt-4o}  & 20.79\% ($\uparrow$ 2.04\%) & 27.47\% ($\uparrow$ 0.95\%) \\
\texttt{o4-mini}  & 49.77\% ($\uparrow$ 1.61\%) & 57.20\% ($\uparrow$ 1.58\%) \\
\texttt{gpt-4o-mini} & 7.93\% ($\downarrow$ 0.35\%) & 12.70\% ($\uparrow$ 0.42\%) \\
\bottomrule
\end{tabularx}
\end{table}

%% file: table/captioning_comparison.tex
\begin{table}
\scriptsize
\centering
\setlength\tabcolsep{8pt}
\caption{\textbf{Comparison of models using caption-only inputs.} Without visual information, particularly GPT-o4-mini shows a huge drop in performance, while smaller models like Qwen2.5-VL even shows a small amount of improvement with detailed captions.}
\label{table:captioning}
\begin{tabularx}{\linewidth}{Xcc}
\toprule
\textbf{Model} & \textbf{Naive} & \textbf{LLM-Judged} \\
\midrule
\texttt{gpt-4o} & 16.01\% ($\downarrow$ 2.74\%) & 22.48\% ($\downarrow$ 4.04\%) \\
\texttt{o4-mini} & 41.09\% ($\downarrow$ \textbf{7.07\%}) & 47.99\% ($\downarrow$ \textbf{7.63\%}) \\
\texttt{gpt-4o-mini} & 7.85\% ($\downarrow$ 0.43\%) & 12.28\% (0.00\%) \\
\texttt{qwen2.5-VL-7B} & 3.23\% ($\uparrow$ 0.22\%) & 4.69\% ($\uparrow$ 0.71\%) \\
\bottomrule
\end{tabularx}
\end{table}

%% file: table/clip_recall.tex
\begin{table}
  \centering
  \scriptsize
  \caption{\textbf{Selected contrastive VLM retrieval performance} on our probe dataset. Full results in \autoref{tab:vqa_retrieval_split_full}.}
  \label{tab:vqa_retrieval_split}
  \begin{tabularx}{\linewidth}{Xccccc}
    \toprule
    Model/Size/Resolution & R@1 & R@5 & P@1 & MRR & NDCG \\
    \midrule
    OpenAI CLIP/B32/224       & 21.1 & 35.4 & 21.1 & 28.4 & 40.1 \\
    OpenCLIP/B16/224       & 25.9 & 40.3 & 25.9 & 33.0 & 44.0 \\
    MobileCLIP/S2/224     & \underline{31.7} & \textbf{49.8} & \underline{31.7} & \textbf{40.0} & \textbf{50.3} \\
    SigLIP/So400m/224    & 28.0 & 43.3 & 28.0 & 35.8 & 46.4 \\
    SigLIP~2/B16/256       & 26.9 & 37.5 & 26.9 & 32.5 & 43.4 \\
    SigLIP~2/L16/256       & 30.3 & 43.1 & 30.3 & 36.8 & 47.2 \\
    SigLIP~2/So400m‑16/256 & 29.2 & 43.5 & 29.2 & 36.1 & 46.6 \\
    SigLIP~2/So400m‑16/384 & 28.9 & 43.3 & 28.9 & 36.1 & 46.6 \\
    SigLIP~2/So400m‑16/512 & 28.5 & 41.9 & 28.5 & 35.4 & 46.0 \\
    SigLIP~2/GOPT16/256    & \textbf{32.2} & 44.0 & \textbf{32.2} & 38.6 & 48.6 \\
    SigLIP~2/GOPT16/384    & 31.2 & 44.9 & 31.2 & 37.9 & 48.0 \\
    Tulip/B16/224       & 28.7 & 42.1 & 28.7 & 35.0 & 45.6 \\
    Tulip/So400m‑14/384 & \underline{31.7} & \underline{47.0} & \underline{31.7} & \underline{39.0} & \underline{49.1} \\
    Tulip/GOPT16/384    & 30.1 & 45.8 & 30.1 & 37.7 & 48.1 \\
    \bottomrule
  \end{tabularx}
\end{table}

%% file: sections/4_conclusion.tex
\section{Conclusion}

Overall, our work confirms that while Vision-Language Models (VLMs) have made strides, they still lack the lateral thinking and nuanced multi-modal abstractions necessary to solve visual puzzles. This short paper raises several important questions for future work, particularly in areas such as understanding negation, interpreting visual metaphors, and moving beyond perceptual tasks to deeper abstract reasoning. Further exploration is also crucial to understand the limits of iterative refinement and to close the awareness versus execution gap in these models, as well as to explore how iterative examination of visual content during reasoning impacts downstream performance, especially given the observed importance of direct visual access for reasoning models. Ultimately, addressing these challenges will be pivotal in developing VLMs that can not only solve visual puzzles but also exhibit more human-like, robust, and generalizable multi-modal understanding across a spectrum of real-world applications.

\section*{Limitations}

While this short paper represents several insights into the capabilities of VLMs in solving rebus puzzles, several limitations should be acknowledged:

\paragraph{Limited Scope/Diversity.} Because the puzzles are all hand-generated and annotated, the dataset that we explore is relatively small (with only 432 samples). This means that exhaustively exploring the full breadth of possible categories is impossible. Further, though the puzzles are carefully annotated, they may still contain biases or overlook certain nuances present in a broader range of puzzles. Moreover, these puzzles are all English-only, and while rebus puzzles exist in other languages \cite{sarti2024non, salvi2016validation, watson1898rebus}, this analysis does not explore those capabilities.

\paragraph{Reliance on closed-source/OpenAI models.} As shown in \autoref{table:rebus_results}, the analysis presented in this paper is somewhat limited by the core performance of most VLMs. The only VLMs that achieve competency at all on these kinds of puzzles are GPT-based models, meaning that we rely heavily on those models in the analysis in \autoref{sec:results}. This limitation means that our findings may not transfer to VLMs in general, and may be somewhat over-specified to OpenAI.

\paragraph{Dataset Specificity and World Knowledge.} Although efforts were made to minimize reliance on domain-specific knowledge, the nature of rebus puzzles can sometimes involve cultural references or idiomatic expressions that might not be universally understood or represented in the models' training data. The comparison with other benchmarks like REBUS, which has a heavy reliance on world knowledge (e.g., Massachusetts towns) in \autoref{app:rebus} highlights this challenge.

\paragraph{Prompt Variance.} While we explore several prompting strategies for eliciting the best performance from VLMs, the optimal prompt may not be one of the ones that we use in this work, and while in \autoref{app:prompt} we explore several variations, and find that there are minimal differences, that may not necessarily be the case. Somewhat mitigating this limitation is our observation in \autoref{sec:results} that in-context examples did not significantly boost models' performance --- suggesting that inherent reasoning capabilities are a more significant bottleneck than prompt understanding for these models.

\paragraph{Potential Risks.} While our analysis offers insight into the strengths and weaknesses of contemporary VLMs on rebus puzzles, several risks should be considered when interpreting these findings. First, as mentioned above, our study is limited to English-language, culturally Western rebus puzzles, which may not capture the diversity or complexity found in other languages or cultural contexts, potentially limiting the generalizability of our conclusions. Additionally, although our probe set is hand-crafted, we cannot fully exclude the possibility that some puzzle types or answer patterns may overlap with data seen during VLM pretraining, which could affect the assessment of true reasoning ability. Finally, there is a risk that readers may overgeneralize our observations beyond the scope of this study. Our results should not be taken as a comprehensive assessment of VLM capabilities, but rather as a focused exploration of their behavior on a specific set of challenging visual-language tasks.

%% file: sections/5_appendix.tex
The appendix consists of the following further discussion:
\begin{itemize}
    \item \autoref{app:code} provides information on the code release, including links to the code bases and datasets used in the project.
    \item \autoref{app:rebus} provides further comparison between our probe dataset with the REBUS and COLUMBUS datasets.
    \item \autoref{app:in-context learning} provides technical details for the in-context learning experiments.
    \item \autoref{app:skill-prompting} provides technical details for the skill-guided prompting experiments.
    \item \autoref{app:multiple-attempts} provides technical details for the iterative refinement experiments.
    \item \autoref{app:caption-only} provides technical details for our caption-based perception experiments.
    \item \autoref{app:image_retrieval} provides technical details for the image-retrieval experiments.
    \item \autoref{app:evaluation} provides details on our Naive and LLM-Judged accuracy computations.
    \item \autoref{app:prompt} explored the impact of the specific prompt on the overall results.
    \item \autoref{app:models} outlines the models used in this paper.
    \item \autoref{app:qualitative examples} provides qualitative examples of model reasoning on our dataset.
    \item \autoref{app:ai_disclosure} details the use of AI in the creation of this manuscript.
\end{itemize}

\section{Code \& Data}
\label{app:code}

We make the code for our analysis available under the MIT license \href{https://github.com/Kyunnilee/visual_puzzles}{here}. We make our annotations freely available under the MIT license, while all original content created by ESL Vault is  licensed with respect to their \href{https://eslvault.com/terms-of-use/}{terms of use}.

\section{Detailed Comparisons with related work}
\label{app:rebus}

\begin{figure*}
    \centering
    \includegraphics[width=\linewidth]{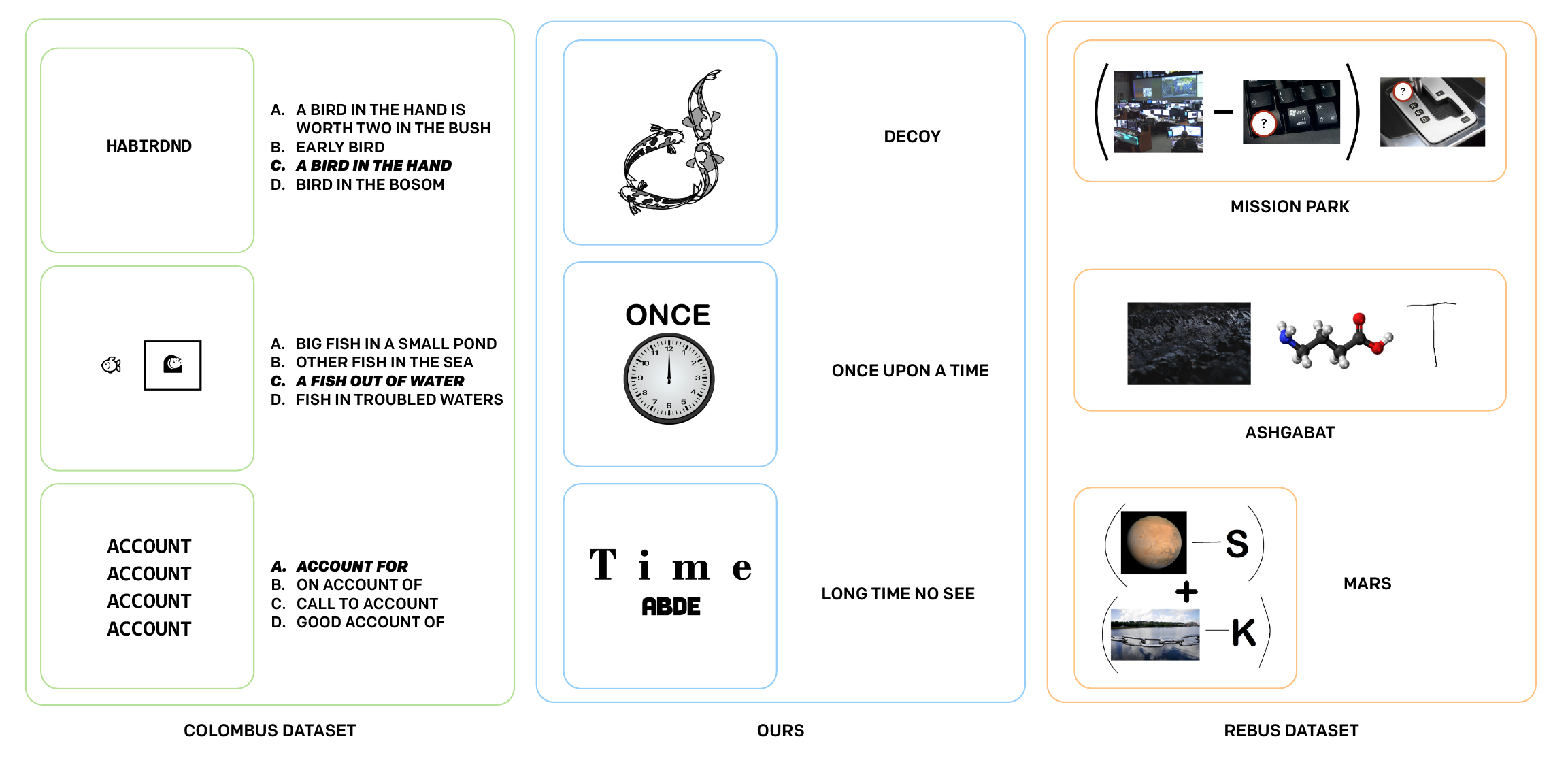}
    \caption{\textbf{Visual comparison between prior work datasets and our probe dataset} (REBUS, COLUMBUS, Ours). COLUMBUS is automatically generated, and uses a limited set of figures and puzzle types. REBUS is hand-generated, but visually inconsistent, and requires a significant amount of world knowledge. Our dataset strikes a balance between the two - requiring challenging lateral thinking and visual understanding skills, but retaining a consistent style and simplified structure. }
    \label{fig:dataset_comparison}
\end{figure*}

\textbf{The REBUS benchmark} \cite{gritsevskiy2024rebus} is a hand-crafted, multi-modal dataset designed to evaluate the reasoning and symbolic understanding abilities of VLMs using 333 original rebus puzzles spanning 13 diverse categories, such as cities, movies, marine life, and more. Each puzzle presents an image-based wordplay challenge that requires models to integrate visual recognition, string manipulation, and multi-step reasoning to infer the correct answer, which is always a single word or phrase matching a specified category. The dataset includes a mix of hand-drawn and digitally composed puzzles and is annotated with advanced characteristics such as inexact spellings, specific real-world references, and the need for reading comprehension. 

The collected puzzles in the REBUS benchmark from \citet{gritsevskiy2024rebus} are primarily focused on real-image understanding. REBUS is heavily biased towards cities, particularly Massachusetts, with 39.64\% of the dataset consisting of city clues, 27.63\% focused on towns in Massachusetts, and 6.01\% consisting of MBTA stations, with only 26.62\% of the data falling outside these categories (into further categories: Marine life ($N=16$, 4.8\%), Composers ($N=14$, 4.2\%), Famous movies ($N=13$, 3.9\%), Surnames ($N=12$, 3.6\%), Food ($N=10$, 3.0\%), Kentucky ($N=7$, 2.1\%), Common phrases ($N=4$, 1.2\%), Christmas songs ($N=3$, 0.9\%), Cartoon characters ($N=3$, 0.9\%), Animals ($N=2$, 0.6\%), Farm things ($N=2$, 0.6\%), Common objects ($N=1$, 0.3\%), Countries ($N=1$, 0.3\%), Weaponry ($N=1$, 0.3\%). 

\textbf{The COLUMBUS benchmark} \cite{kraaijveld2025columbus} is a synthetic, multi-modal dataset designed to evaluate the lateral (creative) thinking capabilities of VLMs through approximately 1,000 multiple-choice rebus puzzles derived from English idioms and compound words. Each puzzle presents either textual or iconographic elements that must be interpreted unconventionally to select the correct phrase or word from four options. The dataset employs a structured, taxonomy-driven methodology to generate puzzles and distractor choices, balancing both visual and linguistic challenges.

The primary limitations of the COLUMBUS benchmark stem from its synthetic design --- each puzzle is generated automatically using a set of 18 latent rules that manipulate visual and textual elements in formulaic ways. These rules are organized into three categories: individual, relational, and modifier. Individual rules alter a single element’s visual properties, such as reversing the order of letters (reverse), changing their direction (up or down), adjusting size (big or small), applying color, crossing out text, or highlighting before, after, or in the middle of a word. Relational rules specify the spatial relationship between elements, such as placing one element next to, inside, above, or outside another. Modifier rules involve repeating an element a specific number of times (e.g., two or four), or substituting an element with a homophone or phonetically similar word (sound). 

This reliance on a fixed taxonomy means that the diversity and creativity of puzzles are inherently limited by the pre-defined rules. This can lead to repetitive puzzle structures and a lack of the nuanced ambiguity, surprise, or cultural references found in human-authored rebus puzzles. Additionally, some rules are underrepresented due to their infrequent occurrence in the source data, resulting in imbalance across the benchmark and possibly biasing models to overfit more common puzzle types. In addition, the distractor choices, generated based on lexical and semantic similarity, may not always be as challenging or as plausible as hand-crafted distractors, sometimes making puzzles either too easy or, conversely, ambiguously difficult. The automated generation also risks producing puzzles with unnatural phrasings or edge cases not typically encountered in real-world lateral thinking scenarios. Finally, the visual presentation is highly controlled, with clean fonts, icons, and spatial arrangements that lack the noise, complexity, and subtlety of real images or handwritten puzzles. This limits the benchmark’s ability to assess a model’s robustness to visual ambiguity and real-world variation.

\autoref{fig:dataset_comparison} shows a visual comparison between the kinds of REBUS puzzles in COLUMBUS, REBUS, and our probe dataset, highlighting some of the details from above.

\section{In-Context Learning}
\label{app:in-context learning}
To investigate whether providing models with in-context examples improves their puzzle-solving performance, we augmented our base prompt with a representative example of a rebus puzzle, including its correct answer and detailed reasoning. We selected an example from cases where models correctly solved the puzzle. The prompt we used for in-context learning is shown below:
\begin{tcolorbox}[promptbox, title={\faComments~ICL Prompt}]
You will be given one example:
<Image>
Response (json):
{
    "answer": "For once in my life",
    "reasoning": "The text 'M1Y L1I1F1E' places the digit '1' between each letter of 'MY LIFE', resulting in four '1's embedded in the phrase. This visually represents 'four 1s in my life', which sounds like 'for once in my life.'",
}
Please solve the rebus puzzle represented by the image. Respond with ONLY a valid JSON object containing two keys:
\begin{enumerate}
\item \texttt{'answer'}: the string value of your solution.
\item \texttt{'reasoning'}: a detailed explanation of how you arrived at this answer, including interpretations of visual elements described and their relationships.
\end{enumerate}
\end{tcolorbox}

\section{Skill-Guided Prompting}
\label{app:skill-prompting}

To systematically analyze whether explicit cognitive skill cues can aid models in solving rebus puzzles, we categorize each puzzle in our dataset according to 11 distinct cognitive skill types, as listed in \autoref{table:annotations}. These skills represent various forms of reasoning required to correctly interpret and solve rebus puzzles.
\input{table/annotations}

\input{table/skill_distribution}
We use a skill-guided prompting strategy, explicitly injecting these annotated cognitive skills into the prompt provided to the model. For each puzzle, we specify the exact skills required, directly informing the model about the cognitive skills it should apply. This is done through the following prompt format:
\begin{tcolorbox}[promptbox, title={\faComments~Skill Guided Prompt}]
<IMAGE> \\
Please solve the rebus puzzle represented by the image. The skills you need to apply are: \{skill\_names\_str\}. Respond with ONLY a valid JSON object containing two keys:
\begin{enumerate}
\item \texttt{'answer'}: the string value of your solution.
\item \texttt{'reasoning'}: a detailed explanation of how you arrived at this answer, including interpretations of visual elements described and their relationships.
\end{enumerate}
\end{tcolorbox}
By injecting these skill annotations into the prompt, we aim to assess whether models can effectively leverage explicit reasoning instructions to improve puzzle-solving performance.
 
\section{Iterative Refinement}
\label{app:multiple-attempts}

To probe the model for its performance in multiple attempts, we run a first initial attempt using the prompt given below:

\begin{tcolorbox}[promptbox, title={\faComments~VLM Iterative Refinement Initial Prompt}]
<IMAGE> \\
This rebus puzzle is a play on words based on images, and may contain text, logical operators, addition/subtraction of letters, and other forms of creative thinking to solve. Can you figure out what it is? Take a deep breath, and let's begin. You can think for as long as you want, until you get a correct answer. When you're done reasoning and thinking, output your final answer as ONLY a valid JSON dict containing the key 'answer' and the string value of your answer. DO NOT include any other text or explanation, such as 'the rebus puzzle represents' or 'the answer is'. Make sure that the JSON dict uses \" instead of ' for the key and value.
\end{tcolorbox}

If the model answers the prompt correctly under the naive matching scheme, we terminate the run. If the model does not answer correctly, we provide the full output of the previous input as context, along with the further user message:

\begin{tcolorbox}[promptbox, title={\faComments~VLM Iterative Refinement Followup Prompt}]
That was incorrect. Your previous answer was '\$previous\_answer'. Please re-examine the image and our conversation history. Take your time, think step by step, and provide a new answer as a JSON dict with the key 'answer'.
\end{tcolorbox}

This iterative process is run for five iterations following the initial prompt (six iterations total), and the overall naive matching accuracy is presented in \autoref{fig:attempts}. We found overall that compliance accuracy was relatively high. With OpenAI, we used the structured decoding API to force the model to provide valid JSON output. For Claude models, we perform significant filtering to isolate JSON outputs, and we found that models failed to provide valid answers in only a maximum of 2/432 examples (where both were refused for safety: ``I apologize, but I do not feel comfortable attempting to solve or interpret that type of puzzle. Perhaps we could have a thoughtful discussion about a different topic that does not involve potentially offensive language or imagery. I'm happy to continue our conversation in a more constructive direction.''). 

\section{Caption-Only Performance}
\label{app:caption-only}

To assess whether the models can accurately solve rebus puzzles without direct visual input, we first generate detailed captions for each puzzle image using GPT-4o, prompted as follows:

\begin{tcolorbox}[promptbox, title={\faComments~Caption Generation Prompt}]
Provide a detailed caption for each image. Do not guess the words or expressions.
\end{tcolorbox}

We then provide only the generated captions and the original prompt to the models, challenging them to solve each rebus puzzle using the following prompt:

\begin{tcolorbox}[promptbox, title={\faComments~Caption-Only Prompt}]
Please solve the rebus puzzle represented by the following caption: \{caption text\}. Respond with ONLY a valid JSON object containing two keys:
\begin{enumerate}
\item \texttt{'answer'}: the string value of your solution.
\item \texttt{'reasoning'}: a detailed explanation of how you arrived at this answer, including interpretations of visual elements described and their relationships.
\end{enumerate}
\end{tcolorbox}

This method enables us to understand the model’s textual reasoning ability, to analyze the impact of visual perception on its overall problem-solving performance.

\section{Image Retrieval}
\label{app:image_retrieval}

In \autoref{sec:results}, we evaluate the performance of CLIP-style VLMs on the probe dataset. Each image $\mathbf{x}_i$ is associated with a corresponding answer $y_i$, forming a dataset $\mathcal{D} = \{(\mathbf{x}_i, y_i)\}_{i=1}^N$. The evaluation protocol is as follows:

\paragraph{Feature Extraction} All images are processed using a pretrained VLM, which encodes each image $\mathbf{x}_i$ into a normalized feature vector $\mathbf{v}_i \in \mathbb{R}^d$. Similarly, all ground-truth answers $\{y_j\}_{j=1}^N$ are tokenized and encoded into normalized text feature vectors $\{\mathbf{t}_j\}_{j=1}^N$.

\paragraph{Similarity Computation} For each image $\mathbf{x}_i$, we compute the similarity between its feature vector $\mathbf{v}_i$ and all text feature vectors $\{\mathbf{t}_j\}$ using the scaled dot product:$$
s_{ij} = \mathrm{softmax}\left( \alpha \cdot \mathbf{v}_i^\top \mathbf{t}_j \right)
$$
where $\alpha$ is a scaling factor (set to 100.0 in our experiments), and the softmax is applied over all $j$ for each $i$.

\paragraph{Ranking and Metrics} For each image, the ground-truth answer $y_i$ is ranked among all candidate answers according to the computed similarities. Let $r_i$ denote the rank of the correct answer for image $i$ (with $r_i = 0$ indicating top-1 retrieval). We report the following metrics:

\vspace{1em}

\noindent \textbf{Recall@K:} The proportion of queries for which the ground-truth answer is ranked within the top $K$ predictions: $$\mathrm{Recall@}K = \frac{1}{N} \sum_{i=1}^N \mathbb{I}[r_i < K]$$
\noindent \textbf{Precision@1:} The fraction of queries where the ground-truth answer is ranked first: $$\mathrm{Precision@1} = \frac{1}{N} \sum_{i=1}^N \mathbb{I}[r_i = 0]$$
\noindent \textbf{Mean Reciprocal Rank (MRR):} $$\mathrm{MRR} = \frac{1}{N} \sum_{i=1}^N \frac{1}{r_i + 1}$$
\noindent \textbf{Normalized Discounted Cumulative Gain (NDCG):} $$\mathrm{NDCG} = \frac{1}{N} \sum_{i=1}^N \frac{1}{\log_2(r_i + 2)}$$

All metrics are computed over the entire evaluation set. The evaluation is performed in a zero-shot manner, with no fine-tuning on the target dataset. Results are reported as aggregate statistics. Full results are given in

\begin{table}
  \centering
  \scriptsize
  \caption{\textbf{Contrastive VLM retrieval performance} on our probe dataset.}
  \label{tab:vqa_retrieval_split_full}
  \begin{tabularx}{\linewidth}{Xccccc}
    \toprule
    Model/Size/Resolution & R@1 & R@5 & P@1 & MRR & NDCG \\
    \midrule
    OpenAI CLIP/B32/224       & 21.1 & 35.4 & 21.1 & 28.4 & 40.1 \\
    OpenCLIP/B16/224       & 25.9 & 40.3 & 25.9 & 33.0 & 44.0 \\
    CoCa/L14/224          & 29.2 & 43.3 & 29.2 & 36.2 & 46.8 \\
    MetaCLIP/L14/224      & 26.6 & 39.6 & 26.6 & 33.3 & 44.1 \\
    MobileCLIP/S2/224     & \underline{31.7} & \textbf{49.8} & \underline{31.7} & \textbf{40.0} & \textbf{50.3} \\
    SigLIP/So400m/224    & 28.0 & 43.3 & 28.0 & 35.8 & 46.4 \\
    SigLIP~2/B16/256       & 26.9 & 37.5 & 26.9 & 32.5 & 43.4 \\
    SigLIP~2/B16/384       & 25.9 & 36.6 & 25.9 & 32.0 & 43.0 \\
    SigLIP~2/B16/512       & 25.0 & 36.8 & 25.0 & 31.3 & 42.3 \\
    SigLIP~2/L16/256       & 30.3 & 43.1 & 30.3 & 36.8 & 47.2 \\
    SigLIP~2/L16/384       & 30.1 & 43.1 & 30.1 & 36.7 & 47.1 \\
    SigLIP~2/L16/512       & 29.6 & 43.1 & 29.6 & 36.5 & 47.0 \\
    SigLIP~2/So400m‑16/256 & 29.2 & 43.5 & 29.2 & 36.1 & 46.6 \\
    SigLIP~2/So400m‑16/384 & 28.9 & 43.3 & 28.9 & 36.1 & 46.6 \\
    SigLIP~2/So400m‑16/512 & 28.5 & 41.9 & 28.5 & 35.4 & 46.0 \\
    SigLIP~2/GOPT16/256    & \textbf{32.2} & 44.0 & \textbf{32.2} & 38.6 & 48.6 \\
    SigLIP~2/GOPT16/384    & 31.2 & 44.9 & 31.2 & 37.9 & 48.0 \\
    Tulip/B16/224       & 28.7 & 42.1 & 28.7 & 35.0 & 45.6 \\
    Tulip/So400m‑14/384 & \underline{31.7} & \underline{47.0} & \underline{31.7} & \underline{39.0} & \underline{49.1} \\
    Tulip/GOPT16/384    & 30.1 & 45.8 & 30.1 & 37.7 & 48.1 \\
    \bottomrule
  \end{tabularx}
\end{table}

\section{Evaluation Method}
\label{app:evaluation}
The evaluation pipeline proceeds in result evaluation then performance estimation via bootstrapping. We conduct two ways for result evaluation.

\paragraph{Naive Matching Evaluation}
In Naive Matching Evaluation, the correctness is determined by directly comparing the predicted answer to the ground truth using exact string matching after normalization (lower casing and stripping whitespace). If the ground truth contains multiple acceptable answers separated by slashes (e.g., apple/orange), the prediction is considered correct if it matches any of the possible normalized answers. Formally, for a predicted answer p and a set of ground truth answers $G = \{g_1, g_2, …, g_n\}$, correctness is evaluated as:
\[
\text{Correct} = 
\begin{cases}
1 & \text{if } p \in G \\
0 & \text{otherwise}
\end{cases}
\]
\paragraph{LLM-Based Semantic Evaluation}
To address limitations of naive matching (e.g., minor spelling errors or paraphrasing), we use an LLM model (GPT-4o) as a judge. The following prompt is provided to the model:

\begin{tcolorbox}[promptbox, title={\faComments~ Default Evaluation}]
\raggedright
Determine whether the predicted answer is semantically equivalent to the ground truth. 
Ignore differences in case and spacing, as well as minor errors in spelling. 
If there's a \texttt{/} in the ground truth, it means that either answer is acceptable.

\textbf{Ground Truth:} \{\{gt\}\} \\
\textbf{Prediction:} \{\{pred\}\} \\
\textbf{Reasoning:} \{\{reasoning\}\} \\

Respond with only \texttt{'yes'} or \texttt{'no'}.
\end{tcolorbox}

The model’s response determines correctness, \texttt{‘yes’} as a correct prediction and \texttt{‘no’} as an incorrect prediction. These evaluation results are recorded in a single JSON log file.

\paragraph{Consistency check for LLM as a judge} To study potential evaluation bias in our LLM-as-a-judge setup, we additionally evaluate with a different LLM (Qwen3-8B with thinking enabled) as a judge. The results are shown in \autoref{table:evaluation_consistency}. The results are comparable to those from GPT-4o, with no change in ranking, indicating that cross model validation is consistent.
\input{table/qwen_as_judge}
\paragraph{Bootstrapped Confidence Estimation}
Using the JSON evaluation log created, we perform repeated sampling with replacement to simulate multiple evaluation scenarios. Specifically, we draw N bootstrap samples of size |S| (equal to the number of total examples), calculate the accuracy for each sample, and estimate the distribution of accuracies.
Formally, for each bootstrap sample S with size |S|, the accuracy is computed as:
\[
\text{Accuracy}_S = \frac{1}{|S|} \sum_{i=1}^{|S|} \mathbb{I}(\text{Correct}_i)
\]

where \( \mathbb{I}(\text{Correct}_i) \) is the indicator function, which returns 1 if the \(i\)-th prediction is correct, and 0 otherwise.

After generating \( N \) bootstrap samples, the 95\% confidence interval is computed.

\paragraph{Details on Human Evaluation Method}
We provide details on the set up of human evaluation experiments. The purpose of these evaluations are (1) highlight the gap between VLMs and expert solvers, and (2) provide a detailed error analysis of the failures in VLMs. We recruited several volunteers from our lab - one “expert” (a native English speaker who regularly solves rebus puzzles) and four “non-experts” (one native and three non-native English speakers with little or no prior exposure). Each participant attempted several puzzles (between 25 and 200) that were randomly drawn from the same evaluation set used for the models. The goal with this experiment was not to perform a rigorous comparison between human and model performance, but to give a general idea of how a small subset of humans perform on this task.

\section{Prompting}
\label{app:prompt}
The prompt used to calculate the overall accuracy in \autoref{table:rebus_results} is provided below:

\begin{tcolorbox}[promptbox, title={\faComments~ Default Evaluation Prompt}]
Please solve the rebus puzzle represented by the image. Respond with ONLY a valid JSON object containing two keys:
\begin{enumerate}
\item \texttt{'answer'}: the string value of your solution.
\item \texttt{'reasoning'}: a detailed explanation of how you arrived at this answer, including interpretations of visual elements described and their relationships.
\end{enumerate}
\end{tcolorbox}

To evaluate the robustness of the prompts, excluding variations such as in-context learning, caption-based prompting, and skill annotation prompts, we tested multiple formulations. Additional evaluated prompts and their corresponding results are listed below.
\begin{tcolorbox}[promptbox, title={\faComments~ REBUS Prompt}]
This rebus puzzle is a play on words based on images, and may contain text, logical operators, addition/subtraction of letters, and other forms of creative thinking to solve. 

Take a deep breath, and let's begin. You can think for as long as you want, until you get a correct answer in the category \$category. When you're done reasoning and thinking, output your final answer with ONLY a valid JSON object containing two keys:

\begin{enumerate}
\item \texttt{'answer'}: the string value of your solution.
\item \texttt{'reasoning'}: a detailed explanation of how you arrived at this answer, including interpretations of visual elements described and their relationships.
\end{enumerate}
\end{tcolorbox}
This prompt is primarily based on the work from \citet{gritsevskiy2024rebus}. We also tested this alternative prompt that provides the general strategies commonly needed to solve rebus puzzles:
\begin{tcolorbox}[promptbox, title={\faComments~ Alternative Prompt}]
This rebus puzzle is a play on words based on images and may contain text, logical operators, addition/subtraction of letters, and other forms of creative thinking to solve.

Here are some tips to solve rebus puzzles:
\begin{enumerate}
    \item Position: Take note of the position of the words and images. 
    \begin{itemize}
        \item Consider prepositions such as \textit{in}, \textit{on}, \textit{above}, \textit{below}, \textit{over}, and \textit{under}.
    \end{itemize}
    \item Size: 
    \begin{itemize}
        \item The size of words and images can provide clues to the answer.
    \end{itemize}
    \item Direction: 
    \begin{itemize}
        \item Pay attention to the direction of words.
        \item Vertical words often include \textit{up} and \textit{down} in the answer.
        \item Backwards words often include \textit{back} in the answer.
    \end{itemize}
    
    \item Style: 
    \begin{itemize}
        \item The style of words and images can also provide important clues.
    \end{itemize}
\end{enumerate}

When you're done reasoning and thinking, output your final answer with ONLY a valid JSON object containing two keys:
\begin{enumerate}
    \item \texttt{'answer'}: the string value of your solution
    \item \texttt{'reasoning'}: a detailed explanation of how you arrived at this answer, including the meaning of each visual element and how they combine
\end{enumerate}
\end{tcolorbox}

The naive matching performance for the three prompts is given in \autoref{tab:app_prompt_perf}, for the GPT-4o model, where we see there is some variance, but it lies well within the 95\% CI for the model given in \autoref{table:rebus_results}.

\begin{table}[h]
    \centering
    \caption{Impact of changing the base prompt on the overall model accuracy (\% success).}
    \label{tab:app_prompt_perf}
    \begin{tabularx}{\linewidth}{Xr}
    \toprule
    \textbf{Prompt} & \textbf{Accuracy} \\
    \midrule
    Default &  18.75 \\
    REBUS & 19.21\\
    Default 2 & 19.68\\
    \bottomrule
    \end{tabularx}
\end{table}

\section{Models}
\label{app:models}
We detail all models used in this project below.

\paragraph{Non-reasoning models} These include models that do not natively support multi-step reasoning with intermediate thoughts or ``thinking tokens.'' GPT-4o and GPT-4o-mini \cite{hurst2024gpt} are widely-used non-reasoning VLMs released by OpenAI in 2024. Gemini 2.0-flash \cite{google2024gemini20} is a similar model released by Google. Claude-3.7-sonnet and Claude-3.5-haiku \cite{TheC3} are non-reasoning variants from Anthropic. Qwen2.5-VL (7B) \cite{bai2025qwen2} and Phi-4 \cite{abdin2024phi} are open-weight, non-reasoning models developed by Qwen and Microsoft, respectively. While the latest Qwen3 \cite{yang2025qwen3} and Phi-4 have reasoning-enabled language-only variants, they are excluded here as they lack multimodal capabilities.

\paragraph{Reasoning models} These models are capable of multi-round chain-of-thought reasoning using special "thinking" tokens, such as OpenAI o3 and o4-mini \cite{openai2025o3} models. Gemini-2.5-flash and Gemini-2.5-pro \cite{google2025gemini25}, both from Google, also support this functionality. Notably, Gemini 2.5-flash has both reasoning and non-reasoning variants and we use the reasoning-enabled version in our experiments.

\paragraph{Decoding configuration} For all models, we set the maximum number of output tokens to a high enough value to avoid truncation and use a temperature of 0 to ensure deterministic decoding. All other settings are left as default.

\section{Qualitative Examples}
\label{app:qualitative examples}

\begin{figure}
    \centering
    \includegraphics[width=\linewidth]{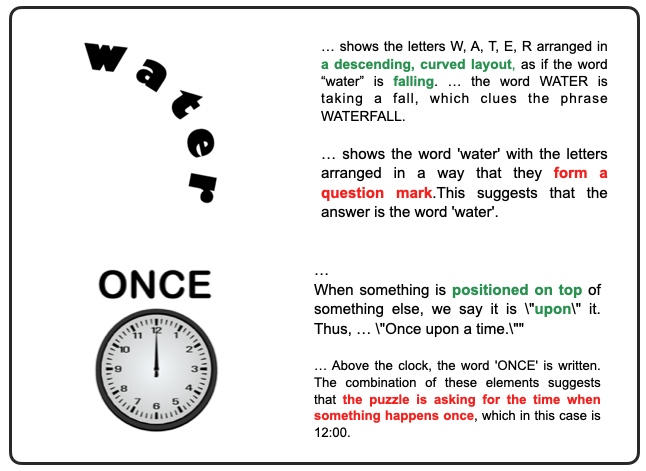}
\caption{This figure shows qualitative examples for both successful (GPT-O3) and failed (Bottom: Qwen 2.5-VL (7B)/Top: Phi-4) cases of model prediction. Green indicates the correct cases, Red is the failure case.}
    \label{fig:qualitative}
\end{figure}

\begin{figure}
    \centering
    \includegraphics[width=\linewidth]{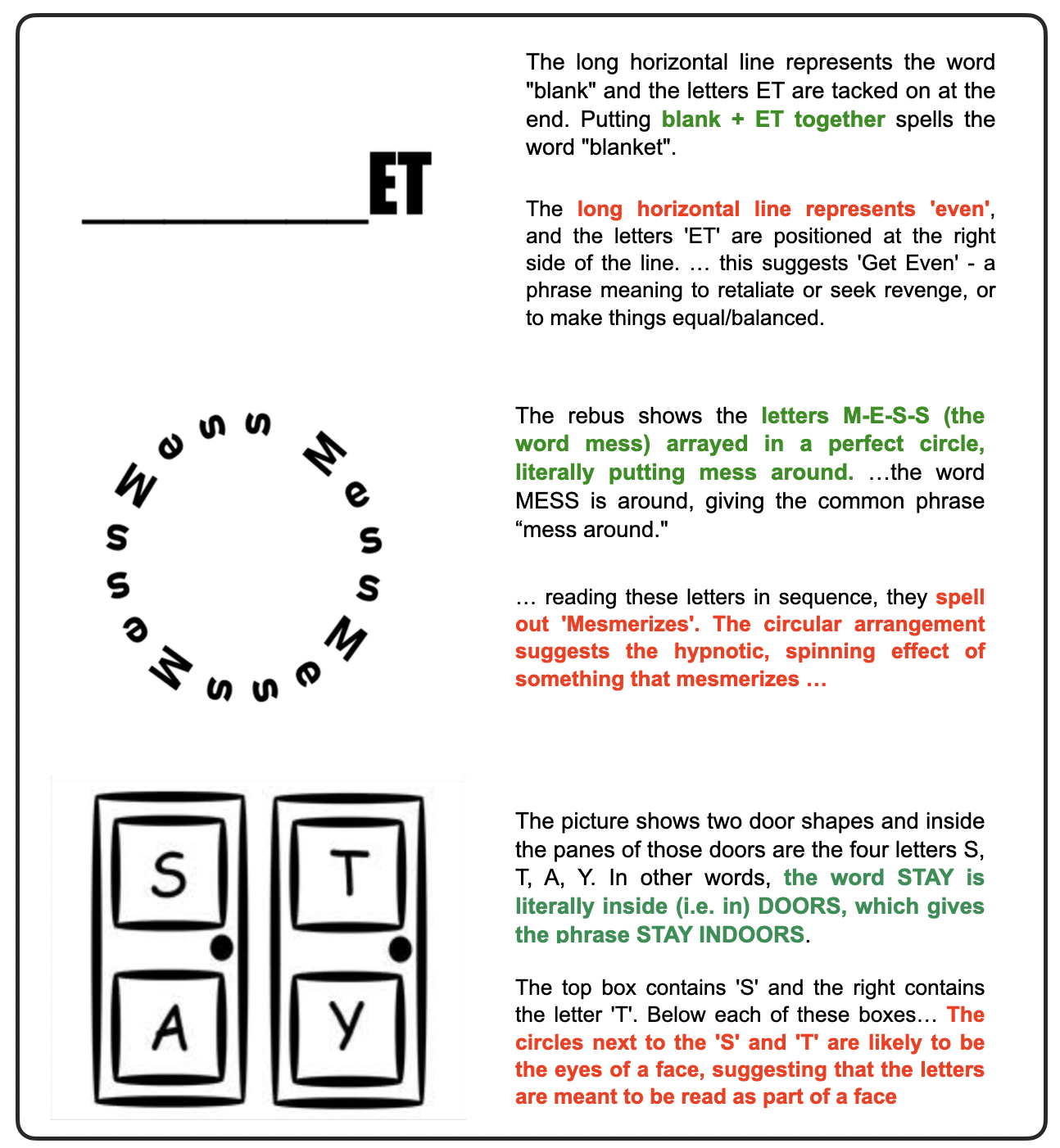}
\caption{This figure shows qualitative examples for both successful (o4-mini) and failed (Claude-3-5-haiku/Last: Phi-4) cases of model prediction.}
    \label{fig:qualitative_2}
\end{figure}

In this section, \autoref{fig:qualitative} shows several qualitative examples from two models (GPT-O3, Phi-4, and Qwen 2.5-VL)and \autoref{fig:qualitative_2} between models (o4-mini, Claude-3-5-haiku, Phi-4) in our probe dataset. 

\section{Disclosure of AI Usage}
\label{app:ai_disclosure}

The authors acknowledge the use of artificial intelligence (AI) tools in the preparation of this manuscript. Specifically, Microsoft Copilot, OpenAI ChatGPT, and Google Gemini Pro were utilized for general editing and code generation / completion purposes. All generated code and text was verified for correctness by one or more of the authors. 

%% file: table/annotations.tex
\begin{table}
\centering
\caption{\textbf{Cognitive skill annotation categories} used for labeling each rebus puzzle in our data set.}
\label{table:annotations}
\begin{tabularx}{\linewidth}{X}
\toprule
\textbf{Category (Abbreviation)}\\
\midrule
\textbf{Image Recognition (IR)} \\
\everypar{\hangindent=1em\hangafter=0}Identifying objects, people, actions, or symbols. \\
\textbf{Text Recognition (OCR + Typography/Layout) (TR)} \\
\everypar{\hangindent=1em\hangafter=0}Detecting words, fonts, capitalization, or stylized text. \\
\textbf{Font Style/Size (FS)} \\
\everypar{\hangindent=1em\hangafter=0}Recognizing different font styles, sizes, or colors. \\
\textbf{Text Orientation (TO)} \\
\everypar{\hangindent=1em\hangafter=0}Understanding text direction (e.g., upside down, rotated) and how it affects meaning. \\
\textbf{Spatial and Positional Reasoning (SPR)} \\
\everypar{\hangindent=1em\hangafter=0}Understanding (multi-)object layout or relationships (e.g., above/below, inside/outside) and how that changes meaning (``man in the moon''). \\
\textbf{Phonetics and Wordplay (PW)} \\
\everypar{\hangindent=1em\hangafter=0}Homophones, puns, mondegreens (``10 issues'' $\rightarrow$ ``tennis shoes''). \\
\textbf{Symbolic Substitution (SS)} \\
\everypar{\hangindent=1em\hangafter=0}Replacing with numbers, letters, or emojis (e.g., ``4'' $\rightarrow$ ``for''). \\
\textbf{Visual Metaphors and Cultural References (VMCR)} \\
\everypar{\hangindent=1em\hangafter=0}Idioms, memes, or visual sayings (``water'' shaped like a ``waterfall''). \\
\textbf{Letter and Word Manipulation (LWM)} \\
\everypar{\hangindent=1em\hangafter=0}Overlapping/hiding/repeating letters to form meanings. \\
\textbf{Absence or Negation (AN)} \\
\everypar{\hangindent=1em\hangafter=0}Missing elements or crossed-out text (e.g., a gap = ``invisible''). \\
\textbf{Quantitative or Mathematical Reasoning (QMR)} \\
\everypar{\hangindent=1em\hangafter=0}Math symbols, object counting (e.g., ``1 2 3'' + foot = ``three feet''). \\
\bottomrule
\end{tabularx}
\end{table}

%% file: table/skill_distribution.tex
\begin{table}
\centering
\small
\setlength\tabcolsep{6pt}
\caption{Distribution of Skill Categories in the Dataset.}
\label{table:skill_distribution}
\begin{tabularx}{\linewidth}{Xr}
\toprule
\textbf{Skill Category} & \textbf{Percentage} \\
\midrule
Absence or Negation & 1.17\%  \\
Font Style/Size & 4.20\%  \\
Image Recognition & 29.42\% \\
Letter and Word Manipulation & 6.06\%  \\
Phonetics and Wordplay & 7.62\%  \\
Quantitative/Mathematical Reasoning & 1.76\%  \\
Spatial and Positional Reasoning & 9.48\%  \\
Symbolic Substitution & 3.42\%  \\
Text Orientation & 1.37\%  \\
Text Recognition (OCR/Layout) & 33.24\% \\
Visual Metaphors and Cultural References & 2.25\% \\
\bottomrule
\end{tabularx}
\end{table}

%% file: table/qwen_as_judge.tex
\begin{table}
\scriptsize
\centering
\setlength\tabcolsep{8pt}
\caption{\textbf{Evaluating LLM-Judge Evaluation Consistency} Qwen3-8B was additionally used as a LLM judge for consistency check.}
\label{table:evaluation_consistency}
\begin{tabularx}{\linewidth}{Xrr}
\toprule
\textbf{Model} & \textbf{GPT-4o LLM-Judge} & \textbf{Qwen3-8B LLM-Judge} \\
\midrule
\texttt{o4-mini} & $55.62_{\textcolor{gray}{[0.509, 0.599]}}$ & $56.60_{\textcolor{gray}{[0.523, 0.613]}}$ \\
\texttt{gemini-2.5-pro} & $47.27_{\textcolor{gray}{[0.424, 0.519]}}$ & $49.54_{\textcolor{gray}{[0.449, 0.542]}}$ \\
\texttt{gpt-4o} & $26.52_{\textcolor{gray}{[0.225, 0.308]}}$ & $27.51_{\textcolor{gray}{[0.231, 0.322]}}$ \\
\texttt{Claude-3.7-Sonnet} & $16.31_{\textcolor{gray}{[0.127, 0.199]}}$ & $19.91_{\textcolor{gray}{[0.162, 0.241]}}$ \\
\bottomrule
\end{tabularx}
\end{table}